\documentclass[acmlarge]{acmart}
\AtBeginDocument{%
  }

\setcopyright{acmlicensed}
\copyrightyear{2018}
\acmYear{2018}
\acmDOI{XXXXXXX.XXXXXXX}

\usepackage{times}  
\usepackage{helvet}  
\usepackage{courier}  
\usepackage{graphicx} 
\usepackage{algorithm}
\usepackage{algorithmic}
\usepackage{newfloat}
\usepackage{listings}
\usepackage{natbib}  
\usepackage{caption} 
\usepackage{bibentry}
\usepackage{tikz}
\usepackage{multirow} 
\usepackage{amsmath}
\usepackage{pgfplots}
\pgfplotsset{compat=1.18}
\usepackage{float}
\usepackage{amsmath}
\usepackage{amsfonts}
\usepackage{algorithm}
\usepackage{algorithmic}
\newcommand{\CALL}[2]{{\scshape #1}(#2)}
\DeclareMathOperator{\avgsim}{avg\,sim}
\DeclareMathOperator*{\avg}{average}

\DeclareCaptionStyle{ruled}{labelfont=normalfont,labelsep=colon,strut=off} 
\floatstyle{ruled}
\newfloat{listing}{tb}{lst}{}
\floatname{listing}{Listing}

\newcommand{\domain}{\mathcal{D}}

\begin{document}

\title{ProToMEx: Rapid, Interpretable Explanations via Structured Representations}

\author{Athina Georgara}\correspondingauthor
\authornote{Both authors contributed equally to this research.}
\email{a.georgara@soton.ac.uk}
\affiliation{%
  \institution{University of Southampton}
  \city{Hampshire}
  \state{England}
  \country{UK}
}

\author{Adarsh Valoor}
\authornotemark[1]
\affiliation{%
  \institution{University of Southampton}
  \city{Hampshire}
  \state{England}
  \country{UK}
}

\author{Sarvapali D. Ramchurn}
\affiliation{%
  \institution{University of Southampton}
  \city{Hampshire}
  \state{England}
  \country{UK}
}

\renewcommand{\shortauthors}{Georgara et al.}

\begin{abstract}
Existing post-hoc explainers for machine learning classifiers primarily focus on feature attribution, assigning importance scores to individual features. While valuable, this approach struggles to articulate the complex, combinatorial patterns that often drive a model's decision-making process. To overcome this limitation, we introduce ProToMEx, a new paradigm for explainability that leverages Probabilistic Topic Models (PTMs). Our model-agnostic framework learns latent ``topics'' that represent distinct, high-level reasons for a classification, moving beyond simple feature importance to reveal underlying semantic structures. ProToMEx naturally provides both global explanations of a model's overall behaviour and local explanations that can disentangle multiple co-existing reasons for a specific prediction. We demonstrate empirically that ProToMEx not only produces explanations of comparable fidelity to popular methods like SHAP and LIME but also drastically reduces the amortised computational cost of generating local explanations, making it highly suitable for real-time applications. Specifically, we show that ProToMEx is $\sim30$-$40\times$ faster than SHAP and LIME over standardised tabular datasets\footnote{The datasets are publicly available in open repositories.} and synthetic datasets.
\end{abstract}




\received{20 February 2007}
\received[revised]{12 March 2009}
\received[accepted]{5 June 2009}

\maketitle
\section{Introduction}

The increasing deployment of complex, black-box machine learning models in high-stakes domains such as finance and healthcare has made the development of faithful post-hoc explanations a critical research imperative. The dominant paradigm in eXplainable AI (XAI) is \emph{feature attribution}, with methods like SHAP \cite{10.5555/3295222.3295230} and LIME \cite{ribeiro2016should} quantifying the contribution of each feature to an individual prediction. While valuable, these methods produce a weighted list of features that, while indicating feature importance, may not constitute a semantically coherent explanation. A list of influential features can be ambiguous or fail to reveal the combinatorial logic underlying a model's decision. As Miller notes, effective explanations are often simple and cite fewer, more cohesive causes \cite{miller2019explanation}. This suggests a fundamental gap: current methods excel at identifying \emph{which} features matter, but often struggle to explain \emph{why} they matter together as a coherent reason.

We contend that a single prediction may be underpinned by multiple, distinct lines of reasoning. For instance, a loan application might be denied due to either a pattern of low income and high debt, \emph{or} a pattern of unstable employment history. A simple feature ranking conflates these distinct reasons. To address this, we introduce a new paradigm for explanation that models these underlying reasons as latent \emph{topics}. We propose \textbf{ProToMEx}, a novel, model-agnostic explainer that leverages Probabilistic Topic Models - a class of generative statistical models from natural language processing---to discover these latent patterns. ProToMEx operates by learning a vocabulary of discretised feature-values and class labels from a classifier's behaviour. It then identifies topics as distributions over this vocabulary, where each topic represents a distinct, interpretable combination of feature values associated with a specific class label. Crucially, the learned topics serve a dual purpose: they act as a global summary of the model's prototypical reasoning patterns across the dataset, while their specific activation in an instance provides a targeted local explanation.

The primary technical challenge in this endeavour is bridging the gap between the structured tabular data used by classifiers and the bag-of-words format required by PTMs. We resolve this with a novel data transformation process that converts instance-label pairs into weighted textual documents. To the best of our knowledge, this is the first work to successfully adapt PTMs for generating post-hoc explanations of any black-box classifier.

Against this background, this paper makes the following contributions to the state-of-the-art:
\begin{enumerate}
    \item We propose \textbf{ProToMEx}, a new model-agnostic explainer that moves beyond feature attribution to \textbf{topic-based reasoning}. It provides both global explanations (prototypical feature patterns for a class) and local explanations (the specific patterns active in a single instance), and is uniquely capable of disentangling multiple reasons for a prediction.

    \item We introduce a \textbf{novel translation methodology} that transforms structured tabular data into a weighted bag-of-words corpus. This technique is the key that unlocks the application of powerful PTMs to the problem of classifier explainability.

\end{enumerate}

The paper is organised as follows. Section~\ref{sec:related_work} reviews related work, Section~\ref{sec:background} provides necessary background, and Section~\ref{sec:protomex} details the ProToMEx framework.
\section{Related work}\label{sec:related_work}

Back et al. in~\cite{bach2015pixel} introduced Layer-wise Relevance BackPropagation (LRP) to explain neural networks (NNs). LRP finds the importance of each feature by backpropagating the score of the output class towards the input layer.
Ribeiro et al. introduced {\em LIME}~\cite{ribeiro2016should}, a model-agnostic method for providing local explanations.
Specifically, given some complex ML model and some input vector, LIME builds a linear model around the input and uses the weights of this local linear model as indicators to identify the most important input features.
Chen and Song~\cite{chen2018learning} propose a framework where a feature selection function is trained to identify the most informative subset of features for each individual data instance.
This function is optimised considering the mutual information between the selected features and the response variable, based on the conditional distribution defined by the model being explained.

Lundberg and Lee put forward {\em SHAP}~\cite{10.5555/3295222.3295230}.
SHAP utilises a game theoretic solution concept, namely the Shapley Values, to compute the marginal contributions of the different input features to the outcome.
SHAP explainer is the current state-of-the-art explainer.
However, computing an explanation with SHAP can be a computationally demanding task, especially as the number of features increases.
This is due to the fact that SHAP needs to compute the Shapley value of each feature, considering all the possible combinations of features---which grows exponentially. 
Notably, even though SHAP acknowledges that the joint contribution of features when combined in different ways varies, the explanations it builds partially convey this information through the features' importance weights. 
Opposed to SHAP, the model we introduce in this paper provides explanations capturing such knowledge.

Beyond feature importance, Ribeiro et al. in~\cite{ribeiro2018anchors} introduced a novel approach to explaining the predictions of complex machine learning models. The authors put forward the concept of `anchors', which are high-precision rules anchoring a prediction locally---as soon as the rules of an anchor are satisfied, the prediction remains consistent.
Radja et al.~\cite{radja2022towards} proposed a novel approach for explaining multi-label classification
models based on formal concept analysis (FCA).
Specifically, the FCA-based explainer identifies properties related to outcomes. 

Different to the works above, ProToMEx identifies patterns of features associated with class labels, and the explanation provided corresponds to a {\em reason}, i.e., a selection of features that together justify the outcome.

\section{Background}\label{sec:background}
Here we provide the necessary background on ML classifiers and Probabilistic Topic Models.
\subsection{ML Classifiers}
Classification refers to the problem of assigning class labels to data points, and more precisely, supervised classification aims to establish a rule that associates a new observation (data point) with one of the existing classes~\cite{michie1995machine}.
An observation is denoted with $x$ and is characterised by $n$ features (we denote the set of all features as $F = \{feature_1,feature_2,\cdots, feature_n\}$).
Let $\domain_i$ be the domain space of feature $i$, i.e., feature $i\in F$ only exhibits values in $\domain_i$ for any observation; thus, observation $x\in \domain$ is an $n$-element vector with $\domain = \domain_1\times\domain_2\times\cdots\times\domain_n$.
Given a set of class labels $\mathcal{L} = \{l_1,\cdots,l_q\}$, we assume the existence of a rule $f : \domain \to \mathcal{L}$ that maps each observation $x \in \domain$ with a label $l\in \mathcal{L}$.
Therefore, an ML classifier, $f_{ML}$, approximates rule $f$ by considering a set of $K$ observations ($X \in \domain^{K}$) along with their associated labels ($Y\in \mathcal{L}^{K}$).

\subsection{Probabilistic Topic Modelling}
Probabilistic Topic Models (PTMs) are a class of unsupervised generative models used to discover latent semantic structures within a collection of documents, known as a corpus~\cite{blei2012probabilistic}. Formally, a corpus $D = \{d_1, \dots, d_m\}$ is a collection of $m$ documents, where each document is a sequence of words from a fixed vocabulary $V$. The core function of a PTM is to uncover the underlying semantic themes, or \textbf{latent topics}, that permeate the corpus. 
Formally, a topic is a multinomial distribution over the vocabulary $V$.

A foundational PTM is \textbf{Latent Dirichlet Allocation (LDA)}~\cite{blei2003latent}. LDA posits a generative process where each document is modelled as a random mixture over a set of $K$ latent topics. The generative process for each document $d$ in the corpus is as follows:
\begin{enumerate}
    \item Draw topic proportions $\theta_d \sim \text{Dir}(\alpha)$.
    \item For each of the $N_d$ words $w_{d,n}$ in the document:
    \begin{enumerate}
        \item Draw a topic assignment $z_{d,n} \sim \text{Categorical}(\theta_d)$.
        \item Draw the observed word $w_{d,n} \sim \text{Categorical}(\phi_{z_{d,n}})$.
    \end{enumerate}
\end{enumerate}
Here, $\alpha$ is the parameter for the Dirichlet prior on the per-document topic distributions, and $\{\phi_k\}_{k=1}^K$ are the topic-word distributions. Given an observed corpus, the primary computational task is to perform posterior inference to estimate the latent variables---the per-document topic mixtures $\{\theta_d\}$ and the shared topic-word distributions $\{\phi_k\}$.


\section{ProToMEx: A Probabilistic Topic Model Explainer}\label{sec:protomex}

We introduce \textbf{ProToMEx}, a novel, model-agnostic framework that leverages Probabilistic Topic Models (PTMs) to generate post-hoc explanations for any black-box classifier. ProToMEx operates by first learning a latent topic space where each topic represents a semantically coherent co-occurrence of feature-values and class labels. These topics are then used to provide both global explanations of a model's overall behaviour and local explanations for individual predictions. A key advantage of our approach is its ability to disentangle and present multiple, distinct reasons for a given classification outcome. The entire process, from model training to explanation generation, is detailed in Algorithm~\ref{alg:phase1_training} and Algorithm~\ref{alg:phase2_explanation}.

The methodology consists of two phases: an offline phase for corpus construction and topic model training, and an online phase for generating explanations for new instances.

\begin{algorithm}[t]\small
\caption{ProToMEx: Phase 1 - Offline Corpus Construction \& Training}
\label{alg:phase1_training}
\begin{algorithmic}[1]
\REQUIRE Black-box classifier $f_{ML}$, feature domain $\mathcal{D}$, number of synthetic samples $m$, number of topics $K$, number of feature-weighting bins $|B|$, discretization factor $\delta$.
\ENSURE A trained PTM represented by topic-word distributions $\Phi$, and the set of class-conditional feature distributions $\{\mu_i^l, \text{idx}_i^l\}$.

\STATE $X \gets \emptyset$ \COMMENT{Initialize synthetic dataset}
\FOR{$j = 1$ to $m$}
    \STATE Sample $\mathbf{x}_j \sim \mathcal{D}$
    \STATE $l_j \gets f_{ML}(\mathbf{x}_j)$
    \STATE $X \gets X \cup \{\langle \mathbf{x}_j, l_j \rangle\}$
\ENDFOR
\FOR{each class $l \in \mathcal{L}$ and each feature $i \in F$}
    \STATE Compute probability mass function $\mu_i^l$ and rank function $\text{idx}_i^l$ from the subset $X_l \subseteq X$ using $\delta$ intervals.
\ENDFOR
\STATE Construct vocabulary $V$ from all feature-value centres and class labels.
\STATE $D \gets \emptyset$ \COMMENT{Initialize corpus}
\FOR{each $\langle \mathbf{x}, l \rangle \in X$}
    \STATE $d \gets $\CALL{TranslateToDocument}{$\mathbf{x}, l, \{\mu_i^l\}, \{\text{idx}_i^l\}, |B|$}
    \STATE $D \gets D \cup \{d\}$
\ENDFOR
\STATE Train LDA on corpus $D$ with $K$ topics to obtain topic-word distributions $\Phi = \{\phi_k\}_{k=1}^K$.
\end{algorithmic}
\end{algorithm}

\begin{algorithm}[t]\small
\caption{ProToMEx: Phase 2 - Online Local Explanation Generation}
\label{alg:phase2_explanation}
\begin{algorithmic}[1]
\REQUIRE A new instance $\mathbf{x}'$, the classifier $f_{ML}$, the trained PTM $\Phi$, and the set of distributions $\{\mu_i^l, \text{idx}_i^l\}$ from Phase 1.
\ENSURE A local explanation for $\mathbf{x}'$: the set of relevant topics $\{k \mid \theta_{d'}(k) > \epsilon\}$ and their corresponding distributions $\{\phi_k\}$.

\STATE $l' \gets f_{ML}(\mathbf{x}')$ \COMMENT{Get the model's prediction for the instance}
\STATE $d' \gets$ \CALL{TranslateToDocument}{$\mathbf{x}', l', \{\mu_i^{l'}\}, \{\text{idx}_i^{l'}\}, |B|$} \COMMENT{Create a document for the instance}
\STATE Infer the topic mixture $\theta_{d'}$ for document $d'$ using the trained model $\Phi$.
\STATE \textbf{return} Set of topics $\{k \mid \theta_{d'}(k) > \epsilon\}$ and their distributions $\{\phi_k\}$.
\end{algorithmic}
\end{algorithm}

\subsection{Offline Phase: Corpus Construction and Topic Modelling}

The core innovation of ProToMEx lies in translating the structured, numerical feature space of a classification problem into a textual corpus suitable for PTMs. This phase is executed once per trained black-box model.

\subsubsection{Synthetic Dataset Generation}
Given a black-box classifier $f_{ML}: \mathcal{D} \to \mathcal{L}$, we first probe its decision boundaries by generating a synthetic dataset $X = \{\langle \mathbf{x}_j, l_j \rangle\}_{j=1}^m$, where each instance $\mathbf{x}_j$ is sampled from the domain $\mathcal{D}$ and its label $l_j = f_{ML}(\mathbf{x}_j)$ is obtained from the classifier. This dataset serves as an empirical representation of the classifier. 

\subsubsection{Document Transformation}\label{subsec:translation}
The central step is the transformation of each instance-label pair $\langle \mathbf{x}, l \rangle \in X$ into a document. This process creates a bag-of-words representation where ``words'' correspond to class labels and discretised feature-value pairs.

\textbf{Vocabulary Construction.} The vocabulary $V$ is the set of all unique words that can appear in our synthetic corpus. It is composed of label-specific tokens and feature-value tokens:
$V = \{\texttt{label\_}\_l\}_{l \in \mathcal{L}} \cup \{\texttt{feat\_}\_i\_\_c \mid i \in F, c \in \bigcup_{l \in \mathcal{L}} \Sigma_i^l\}$, where $F$ is the set of features and $\Sigma_i^l$ is the set of discretized value-centres for feature $i$ conditioned on class $l$.

\textbf{Feature Discretisation and Weighting.} To capture the relationship between feature values and class labels, we first compute class-conditional probability mass functions for each feature. For each class $l \in \mathcal{L}$ and each feature $i \in F$, we discretize the observed value range $[\min_{\mathbf{x} \in X_l} x_i, \max_{\mathbf{x} \in X_l} x_i]$ into $\delta$ uniform intervals, where $X_l \subseteq X$ contains all instances classified as $l$. Let $\Sigma_i^l = \{c_1, \dots, c_\delta\}$ be the set of interval centres. The empirical probability mass function $\mu_i^l : \Sigma_i^l \to [0,1]$ is computed as:
$$
\mu_i^l(c) = \frac{1}{|X_l|} \left| \left\{ \mathbf{x}_j \in X_l \text{ s.t. } x_{j,i} \text{ falls in the interval of } c \right\} \right|
$$
This function, $\mu_i^l$, quantifies the prevalence of a feature's value range for a specific class. We then define a rank function $\text{idx}_i^l: \Sigma_i^l \to \{1, \dots, \delta\}$ that orders the interval centres based on their mass, with higher ranks corresponding to higher probability mass.

\textbf{Document Assembly.} For a given pair $\langle \mathbf{x}, l \rangle$, we construct a document $d$ by populating it with words from $V$. For each feature $i$, we identify the centre $c_{x_i} \in \Sigma_i^l$ corresponding to the value $x_i$. The features are then partitioned into $|B|$ bins based on their rank $\text{idx}_i^l(c_{x_i})$. A feature $i$ in bin $b_k$ (for $k \in \{1, \dots, |B|\}$) results in the word $\texttt{feat\_\_i\_\_}c_{x_i}$ being repeated $k$ times in the document $d$. This weighting scheme ensures that feature-values more strongly characteristic of class $l$ have a greater influence in the document. Finally, the word $\texttt{label\_\_}l$ is added to the document, repeated $|B|$ times to assert its importance.

\subsubsection{Topic Modelling}
With the full corpus $D = \{d_j\}_{j=1}^m$ constructed, we train a standard Latent Dirichlet Allocation (LDA) model \cite{blei2003latent}. The LDA training yields $|K|$ topics, where each topic $k$ is a probability distribution $\phi_k$ over the vocabulary $V$.

\subsection{Online Phase: Explanation Generation}
Once the PTM is trained, generating explanations for the classifier $f_{ML}$ is computationally efficient.

\subsubsection{Local Explanations}
To explain a prediction for a new instance $\mathbf{x}'$, we first obtain its label $l' = f_{ML}(\mathbf{x}')$. We then transform the pair $\langle \mathbf{x}', l' \rangle$ into a document $d'$ using the pre-computed distributions ($\mu_i^{l'}$ and $\text{idx}_i^{l'}$) from the offline phase. Using the trained LDA model, we infer the topic-document distribution $\theta_{d'}$, which represents the mixture of topics that constitute document $d'$.

The local explanation for $\mathbf{x}'$ is the set of topics with non-negligible probability in this mixture, $\{k \mid \theta_{d'}(k) > \epsilon\}$. Each such topic $k$ highlights a distinct combination of feature-values (via its distribution $\phi_k$) that justifies the classification $l'$. This allows ProToMEx to reveal if a prediction is supported by multiple, independent lines of reasoning within the data.

\subsubsection{Global Explanations}
Global explanations describe the general behaviour of the classifier for an entire class, without reference to a specific instance. A global explanation for a class $l \in \mathcal{L}$ is the set of topics that are strongly associated with its corresponding vocabulary word. Formally, this is the set of topics $\{k \mid \phi_k(\texttt{label\_\_}l) > \epsilon'\}$. Each of these topics represents a prototypical instance profile for that class, identifying a distinct pattern of feature-values that the model $f_{ML}$ has learned to associate with the label $l$.

\vspace{-2mm}
\section{Empirical Evaluation}\label{sec:evaluation}

To validate our proposed model, we conducted an empirical evaluation designed to assess \textbf{ProToMEx} on two primary axes: explanation fidelity and computational performance. We benchmark our method against two widely-used explainers, SHAP~\cite{10.5555/3295222.3295230} and LIME~\cite{ribeiro2016should}. The evaluation spans multiple public and synthetic datasets with diverse characteristics, including both binary and multi-class classification tasks, to demonstrate the general applicability of our approach.

\paragraph{Implementation Details.}
Our experiments were coded in Python 3.12. The black-box models to be explained are Neural Networks implemented with \texttt{scikit-learn}~\cite{sklearn_api}. ProToMEx's topic modelling component is built upon the \texttt{Gensim} library 's~\cite {rehurek2011gensim} implementation of LDA. For comparison, we used the official SHAP library (v0.46.0).
The code will be publicly available upon acceptance.

\subsection{Datasets}
In this empirical evaluation, we used both synthetic and publicly available datasets, which we describe below. 
The description is shown in Table~\ref{tab:synthetic_datasets} and Table~\ref{tab:datasets}.
 All datasets can be found in {\url{https://bit.ly/protomex-datasets}}.

\subsubsection{Synthetic Datasets}
We constructed synthetic tabular datasets containing both numerical and categorical data, with a varying number of features and a varying number of class labels. For each dataset, we generated {\em patterns} of features related to each class label.
That is, each feature $f\in F$ can take values in the domain space $\domain_f$, a pattern $\pi$ is characterised by a subset of features $\tilde{F}\subseteq F$ with each feature $f \in \tilde{F}$ taking values from a subdomain space $\tilde{\domain}_f \subseteq \domain_f$ and a single class label $l \in \mathcal{L}$.
For each dataset, we randomly generated features with each feature being numerical with probability 75\%, otherwise categorical. The domain space of categorical data contains 2 to 4 distinct values (randomly selected), while the domain space of all numerical data is $[0,1]$. Given the features $F$ of a dataset, for label $l\in\mathcal{L}$  we generate patterns so that for each pattern $\pi$ it holds:
\begin{enumerate}
    \item the size of the pattern randomly selected from the range $\big[\lceil 0.5\cdot |F|\rceil - 1, \lceil 0.5\cdot |F|\rceil + 1\big]$
    \item the subdomain space $\tilde{D}_f$ for each $f \in \tilde{F}_\pi$ is
    \begin{itemize}
        \item a single value $v \in D_f$ if f is categorical, or
        \item a range in $\big[[0,0.2]$,$[0.2,0.4]$,$[0.4,0.6]$,$[0.6,0.8],[0.8,1]\big]$ if f is numerical
    \end{itemize}
    \item the similarity of pattern $\pi$ with any other pattern $\pi'$ of the dataset is less than $0.35$.
\end{enumerate}
For measuring the similarity between two patterns $\pi$ and $\pi'$ we put forward the similarity metric below:
\begin{equation}\label{eq:similarity}
    sim(\pi,\pi') =  \eta(\alpha_{\pi,\pi'},\beta_{\pi,\pi'}) \cdot 
    \avgsim (\tilde{F}_{\pi}||\tilde{F}_{\pi'})
\end{equation}
with $\avgsim (\tilde{F}_{\pi}||\tilde{F}_{\pi'})$ denoting the average of maximum similarity of each feature $f\in\tilde{F}_{\pi}$ to a feature $f'\in\tilde{F}_{\pi'}$, i.e.,:
$$
    \avgsim(\tilde{F}_{\pi}||\tilde{F}_{\pi'}) = \avg_{f \in \tilde{F}_\pi} \max_{f' \in \tilde{F}_{\pi'}} sim(f,f')
$$
and $sim(f,f') = \frac{\domain_{f} \cap \domain_{f'}}{\domain{f}\cup \domain_{f'}}$ if $f$ and $f'$ refer to the same feature in $F$; otherwise $sim(f,f') = 0$.
Scaling factor $\eta(\alpha_{\pi,\pi'},\beta_{\pi,\pi'})$, inspired by the similarity metric defined in~\cite{li2003approach}, takes into consideration the {\em coverage} and {\em alignment} degree to determine the maximum possible similarity between the two patterns:
\small
$$
    \eta(\alpha_{\pi,\pi'},\beta_{\pi,\pi'}) = \gamma\cdot e^{\lambda_1 \cdot(1-\beta_{\pi,\pi'})} \cdot \frac{e^{\lambda_2\cdot\alpha_{\pi,\pi'}} - e^{-\lambda_2\cdot\alpha_{\pi,\pi'}}}
    {e^{\lambda_2\cdot\alpha_{\pi,\pi'}} + e^{-\lambda_2\cdot\alpha_{\pi,\pi'}}}
$$
\normalsize
with coverage degree $\alpha_{\pi,\pi'} = \frac{\tilde{F}_\pi \cap \tilde{F}_{\pi'}}{\tilde{F}_\pi \cup \tilde{F}_{\pi'}}$ measuring the common features over all features in the patterns, and alignment degree $\beta_{\pi,\pi'} = \frac{\tilde{F}_\pi \cap \tilde{F}_{\pi'}}{\min\{|\tilde{F}_\pi|,|\tilde{F}_{\pi'}|\}}$ measuring the distance of the smallest pattern to become subset of the largest. Finally, 
$\gamma = 1 / \big(e^{\lambda_1}\cdot\frac{e^{\lambda_2}-e^{-\lambda_2}}{e^{\lambda_2}+e^{-\lambda_2}}\big)$ 
is a normalisation factor, and $\lambda_1,\lambda_2 \in \mathbb{R}_+$ are regulating parameters.
 In our experiments, we set $\lambda_1 = 0.1$ and $\lambda_2=0.9$.
 Intuitively, the larger the coverage is, the more similar the two patterns are allowed to be. Similarly, the larger the alignment is (i.e., the closer one of the two patterns is to a subset of the other), the more similar the two patterns are allowed to be.

 We generated a number of data points (varying across the settings) per label by sampling over the patterns related to the label at hand. Intuitively, the pattern used to generate a data point corresponds to {\em reason why} the data point is labelled with a specific class label; therefore, the pattern is a sufficient explanation.


\subsubsection{Loan Approval Dataset}
This dataset contains data regarding financial risk for loan approval~\cite{dataset_loan}.
It consists of $45000$ data points, and each data point is characterised by $13$ features (both categorical and continuous ones).
Specifically, there are $5$ categorical features, $8$ continuous features ($6$ float numbers and $2$ integer numbers).
There are two (2) classes in this dataset: loan approved ($\sim22.23\%$ of the data points) or loan rejected ($\sim77.77\%$ of the data points).
\subsubsection{Lung Cancer Dataset}
This dataset contains data related to various factors that may influence the risk of lung cancer~\cite{dataset_lung}.
 It consists of $3000$ data points, and each data point is characterised by $15$ features ($14$ categorical features and $1$ continuous feature).
 There are two (2) classes in this dataset, healthy lungs ($49.4\%$ of the data points) or malignant lungs ($50.6\%$ of the data points).

\subsubsection{Wine Quality Dataset}
This dataset contains data regarding wine quality~\cite{dataset_wine}.
 It consists of $21000$ data points, and each data point is characterised by $11$ features (all features are continuous).
 There are two (3) classes in this dataset, low quality ($28.57\%$ of the data points), mediocre quality ($42.86\%$ of the data points) and high quality ($28.57\%$ of the data points).

\subsubsection{Air Quality Dataset}
This dataset contains data regarding the air quality across various regions~\cite{dataset_air}.
It consists of $5000$ data points, and each data point is characterised by $9$ features (all features are continuous).
There are four (4) classes in the dataset, air quality: good ($40\%$ of the data points), moderate ($30\%$ of the data points), poor ($20\%$ of the data points) and hazardous ($10\%$ of the data points).

\begin{table}[]
    \centering
    \begin{tabular}{l|c|c|c|c}
         Setting & \# Features & \# Labels &\# Patterns &Size\\\hline
         F8P10L2 & 8 & 2 &10 &20000\\
         F8P6L3 & 8 & 3 &6 &20000\\
         F9P10L2 & 9 & 2 &10 &30000\\
         F9P6L3 & 9 & 3 &6 &30000\\
         F10P4L2 & 10 & 2 &4 &20000\\
         F10P10L2 & 10 & 2 &10 &20000\\
         F10P15L3 & 10 & 3 &15 &30000\\
         F10P20L2 & 10 & 2 &20 &20000\\
         F15P10L2 & 15 & 2 &10 &20000\\
    \end{tabular}
    \caption{Synthetic datasets description 
    We used $60\%$ of the dataset for training the classifier, $35\%$ for training ProToMEx and $5\%$ for testing.
    }
    \label{tab:synthetic_datasets}
\end{table}

\begin{table}
    \centering
    \begin{tabular}{l|c|c|c|c|c}
    \multirow{2}{*}{Dataset}&\multicolumn{3}{c|}{Size (\# data points)}&\multirow{2}{*}{\#Feat}&\multirow{2}{*}{Labels}\\\cline{2-4}
    &Training & Corpus & Testing & \\\hline
         Loan& 31751 & 12349 &900 & 14 & Binary  \\
         Lungs & 2116 & 824 & 60 & 15 & Binary\\
         Wine & 14816 & 5767 & 420 & 11 & Multiple\\
         Air  & 3526 & 1374 &100 & 10 & Multiple\\
    \end{tabular}
    \caption{Publicly available datasets description}
    \label{tab:datasets}
\end{table}

\begin{table}
    \centering
    \begin{tabular}{l|c|c|c|c}
        \multirow{2}{*}{Setting} & Corpus& ProToMEx &
        \multicolumn{2}{c}{ Explain Instance ($\times 10^{-2}$ sec)}\\\cline{4-5}
        & Gen (sec)&training (sec)& SHAP  &ProToMEx\\\hline
         F8P10L2 & 1.53 &54.97&0.46 &0.12\\
         F8P6L3 & 2.33& 86.14& 0.97&0.13\\
         F9P10L2 & 1.59 & 62.09&0.97&0.14\\
         F9P6L3 & 2.32& 90.72& 1.77&0.14\\
         F10P4L2 &1.72 &59.91&3.87 &0.15\\
         F10P10L2 & 1.59 &57.49& 1.83&0.13\\
         F10P15L3 & 2.46 & 82.19 &3.33 & 0.14\\
         F10P20L2 & 1.53 &57.06& 1.78&0.13\\
         F15P10L2 & 1.93& 66.76& 3.05&0.14
    \end{tabular}
    \caption{Synthetic Datasets. Average time required for ProToMEx vs SHAP}
    \label{tab:synthetic:time}
\end{table}
\begin{table}
    \centering
    \begin{tabular}{l|c||c|c|c}
        \multirow{2}{*}{Setting} & Detected& Most Probable& Most Similar& Weighted Avg\\
        & Topics & Relevant Topic & Relevant Topic& of Relevant Topics
        \\\hline
         F8P10L2 &0.81 & 0.75 &0.78& 0.66\\
         F8P6L3 &0.77 & 0.71 &0.74& 0.69\\
         F9P10L2 &0.75&0.7&0.71&0.61\\
         F9P6L3 &0.77&0.66&0.73&0.66\\
         F10P4L2 & 0.82 & 0.71& 0.76&0.68\\
         F10P10L2 &0.66&0.59&0.62&0.49\\
         F10P15L3 &0.72&0.65&0.69&0.56\\
         F10P20L2 &0.62&0.56&0.59&0.43\\
         F15P10L2 &0.79&0.65&0.68&0.56
    \end{tabular}
    \caption{Similarity between topics and original patterns}
    \label{tab:synthetic:similarity}
\end{table}

\subsection{Setup}
Our evaluation protocol is executed independently for each dataset. For each dataset, we partition the data into three disjoint sets: a \textbf{training set} (60\% for synthetic data; 70\% for public data), a \textbf{corpus set} (35\% and 28\%, respectively), and a \textbf{testing set} (5\% and 2\%).
First, we train a black-box classifier, a 5-layer neural network with ReLU activation on the training set. Next, to prepare our explainer, we use the trained classifier to generate predictions for the corpus set. These instance-label pairs are used to construct the document corpus and train the LDA model for ProToMEx, as detailed in Section~\ref{subsec:translation}. Finally, we generate local explanations for each instance in the testing set using ProToMEx, SHAP, and LIME. To ensure statistical robustness, we repeat this entire procedure 100 times with different random seeds. All reported results are the average values across these runs.

\subsection{Results}

We analyse the computational performance of ProToMEx against SHAP on synthetic datasets, with results averaged over 100 runs shown in Table~\ref{tab:synthetic:time} (Fig.~\ref{fig:synthetic:shape_vs_protomex}). The performance of ProToMEx is characterised by two phases. First, a one-time offline \emph{overhead} is incurred for corpus generation and LDA model training, a cost that scales with corpus size. Second, during the online phase of explaining an individual instance, ProToMEx is up to \textbf{25$\times$ faster than SHAP}. This significant speed-up is because ProToMEx utilises its pre-trained model for inference, whereas SHAP must recompute its explanation for every new instance. As we show later in this section, this performance advantage over competing methods is even greater on public datasets.

Next, we evaluate the capability of ProToMEx to identify the reasons why an instance is classified with a specific class label. We constructed the synthetic dataset so that each instance is created using some pattern, with the pattern denoting the reasons for labelling the instance in a specific way.
Our evaluation explores two aspects.
First, we assess whether ProToMEx extract topics that match to patterns used to create a given dataset.
Second, we assess how well ProToMEx can identify for each instance the specific pattern it was generated with.
Table~\ref{tab:synthetic:similarity} shows the corresponding results.
For measuring the similarity between the patterns and the topics detected by ProToMEx, we use the similarity metric described in Equation~\ref{eq:similarity}.
That is, a detected topic is converted into a pattern considering all the non-zero terms. The subdomain domain space of a feature in a topic is the interval $\delta$ reconstructed by the centre used in the term during the document generation (see ``Offline Phase'' in Section~\ref{sec:protomex}).
Notably, the similarity metric entirely neglects the label associated with the pattern. 
As such, we measure the similarity between  a topic ($\pi_\text{topic}$) and a pattern  ($\pi_\text{pattern}$) as $\xi(l_\text{topic}, l_\text{pattern}) \cdot sim(\pi_\text{topic},\pi_\text{pattern})$ where
$\xi(l_\text{topic}, l_\text{pattern}) = 1$ if and only if $l_\text{topic} = l_\text{pattern}$, and $0$ otherwise.
Intuitively, $\xi(l_\text{topic}, l_\text{pattern})$ denotes that two patterns are considered completely dissimilar when their labels are different, i.e., patterns associated with different labels cannot be similar.

In Table~\ref{tab:synthetic:similarity}, we see that ProToMEx extract topics that are very similar (with similarity at least $0.62$ on average) to the patterns with which the dataset at hand was generated. 
Notably, patterns and topics are of different sizes; therefore the similarity drops due to the coverage degree ($\alpha_{\pi,\pi'}$); however the alignment degree ($\beta_{\pi,\pi'}$) of the patterns and the average maximum similarity ($\avgsim)$ increase the similarity between the patterns. 
That is, the original patterns tend to be subsets of the topics discovered by ProToMEx.
As expected, the quality of the topics depends on the number of features of the dataset and the number of patterns used. 

To assess the capability of our model to identify the correct pattern when explaining a single instance, we consider three quantities:
\begin{itemize}
    \item the similarity of the instance's ground truth pattern with the instance's {\em most probable} relevant topic;
    \item the similarity of the instance's ground truth pattern with its most similar across the instance's relevant topics; and
    \item the average over the similarity of the instance's ground truth pattern with any of instance's relevant topics, weighted with the topic's probability.
\end{itemize}
As we can see, most probable relevant topics achieve good similarity with the ground truth, indicating that ProToMEx can provide a close-to-ground-truth reason why an instance is classified with a specific label.
Interestingly, there is a divergence between the most probable and the most similar topics (with most similar exhibiting higher similarity), indicating that ProToMEx manages to associate instances with topics close to the ground truth, but with not quite accurate probability.

Table~\ref{tab:time} (Fig.~\ref{fig:public:lime_vs_shape_vs_protomex}) details the computational requirements of ProToMEx against SHAP and LIME in the publicly available datasets, averaged over 100 runs. Again, the performance of ProToMEx involves two distinct phases: the one-time overhead corpus generation and LDA model training, and the time required to explain a single instance. In this online phase, ProToMEx demonstrates a significant performance advantage, generating explanations at least \textbf{30-40x faster than SHAP} and \textbf{20x LIME}. This is because ProToMEx leverages its pre-trained model, while the baselines must recompute for every instance. Consequently, for applications requiring repeated explanations, the initial time investment is quickly offset. We find that ProToMEx becomes the most time-efficient method after explaining 2600 instances on the Loan dataset, 1225 on Wine, 125 on Air, and just 90 instances on the Lung Cancer dataset.

\begin{table}[h]
    \centering
    \begin{tabular}{l|c|c|c|c|c}
        \multirow{2}{*}{Dataset} & Corpus& ProToMEx &
        \multicolumn{3}{c}{ Explain Instance ($\times 10^{-2}$ sec)}\\\cline{4-6}
        & Gen (sec)&training (sec)& SHAP  & LIME&ProToMEx\\\hline
         Loan & 1.68 &117.92  &1320.04  &3.74 &0.125\\
         Lungs &  0.07 & 6.001 & 94.3 &2.29 &0.121 \\
         Wine &  1.56&26.71  & 484.6 &2.46 &0.106\\
         Air & 0.3& 5.933 & 121.6 &2.76& 0.095
    \end{tabular}
    \caption{Publicly Available Datasets. Average time required for ProToMEx vs SHAP vs LIME.}
    \label{tab:time}
\end{table}

\begin{table*}[h]
\centering
    \begin{tabular}{c|c|c|c|c|c|c|c|c|c|c|c|c|c}
        \multirow{2}{*}{Dataset} & \multicolumn{2}{c|}{\# Influential Features}&
        \multicolumn{10}{c}{\# Instances with \# Missing Influential Features Equal to ($=$)}\\\cline{2-3}\cline{4-14}
        & SHAP  & ProToMEx & 0 & 1& 2& 3& 4& 5& 6& 7& 8& 9& 10\\\hline
        Loan &  5.27 & 9.63&
        332.83	&283.58	&164.5	&37.92	&13.33&	63.5&	4.33&	0&	0&0&0
\\

        Lungs & 10.203&	10.234&
        
        0.36&	4.44&	6.48&	8.12&	14.88&	16.2&	9.04&	0.44&	0.04	&0&	0\\

        Wine &4.12&	8.87&67.2 &
        76.8	&118.1&	114.6&	36&	7.3&	0&	0	&0	&0	&0 \\

        Air & 5.73&	7.21 & 5.73&	33.08&	37.14&	20.44&	3.36&	0.25&	0&	0&	0&-&-
    \end{tabular}
    \caption{Average number of influential features according to SHAP and average number of influential features found by ProToMEx (across the whole testing dataset). Average number of instances (within testing dataset) that ProToMEx missed $x$ influential features. }
    \label{tab:quality:SHAP}
    
\end{table*}
\begin{table*}[h]
\centering
    \begin{tabular}{c|c|c|c|c|c|c|c|c|c|c|c|c|c}
        \multirow{2}{*}{Dataset} & \multicolumn{2}{c|}{\# Influential Features}&
        \multicolumn{10}{c}{\# Instances with \# Missing Influential Features Equal to ($=$)}\\\cline{2-3}\cline{4-14}
        & LIME  & ProToMEx & 0 & 1& 2& 3& 4& 5& 6& 7& 8& 9& 10\\\hline
        Loan &  5.39 & 9.63&
        261.17	&246	&238.67	&16.33	&34.83&	3&	0&	0&	0&0&0 

\\
       Lungs & 8.56&	10.23&
        0.72	&4.32&	13.48&	20.2&	15.48&	5.04	&0.68	&0.08&0&0&0\\

        Wine &4.12&	55.4&68.6 &
        115.3	&130.2&	43.7&	6.8&	0&	0&	0	&0	&0	&0 \\

        Air & 5.67&	7.21 & 4.87&	31.87&	36.69&	23.3&	3.01&	0.26&	0&	0&	0&-&-

    \end{tabular}
    \caption{Average number of influential features according to LIME and average number of influential features found by ProToMEx (across the whole testing dataset). Average number of instances (within testing dataset) that ProToMEx missed $x$ influential features. }
    \label{tab:quality:LIME}
\end{table*}

\begin{figure}
    \centering
\begin{tikzpicture}
\begin{axis}[
    xbar,
    bar width=5pt,
    xlabel={Time ($\times 10^{-2}$ sec)},
    ylabel={Dataset},
    symbolic y coords={
        F8P10L2, F8P6L3, F9P10L2, F9P6L3,
        F10P4L2, F10P10L2, F10P15L3, F10P20L2,
        F15P10L2
        },  
    legend style={at={(1,0)}, anchor=south east},
    legend image code/.code={%
        \draw[#1, yshift=-0.25em] (0cm,0cm) rectangle (0.3cm,0.3em);
    },
    ytick=data,
    yticklabel style={font=\scriptsize},
    nodes near coords,
    xmin=0,
    xmax=4.5,
]
\addplot coordinates {
    (0.12,F8P10L2)
    (0.13,F8P6L3)
    (0.14,F9P10L2)
    (0.14,F9P6L3)
    (0.15,F10P4L2)
    (0.13,F10P10L2)
    (0.14,F10P15L3)
    (0.13,F10P20L2)
    (0.14,F15P10L2)
};
\addlegendentry{ProToMEx}
\addplot coordinates {
    (0.46,F8P10L2)
    (0.97,F8P6L3)
    (0.97,F9P10L2)
    (1.77,F9P6L3)
    (3.87,F10P4L2)
    (1.83,F10P10L2)
    (3.33,F10P15L3)
    (1.78,F10P20L2)
    (3.05,F15P10L2)
};\addlegendentry{SHAP}
\end{axis}
\end{tikzpicture}
    \caption{Explaining single instance in Synthetic Datasets: SHAP vs ProToMEx.}
    \label{fig:synthetic:shape_vs_protomex}
\end{figure}
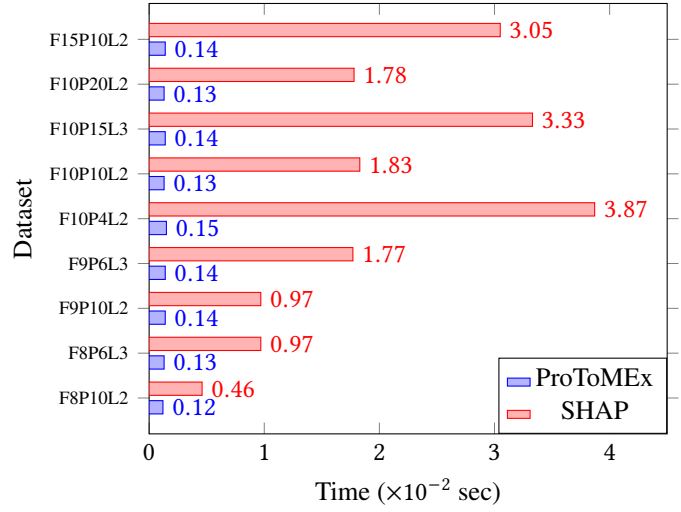
\begin{figure}
    \centering
\begin{tikzpicture}
\begin{axis}[
    xbar,
    bar width=5pt,
    xlabel={Time ($\times 10^{-2}$ sec)},
    ylabel={Dataset},
    symbolic y coords={
        Loan, Lungs, Wine, Air
        },  
    legend style={at={(1,1)}, anchor=north east},
    legend image code/.code={%
        \draw[#1, yshift=-0.25em] (0cm,0cm) rectangle (0.3cm,0.3em);
    },
    ytick=data,
    nodes near coords,
    xmin=0,
]
\addplot coordinates {
    (0.125,Loan)
    (0.121,Lungs)
    (0.106,Wine)
    (0.095,Air)
};
\addlegendentry{ProToMEx}
\addplot coordinates {
    (1320.04,Loan)
    (94.3,Lungs)
    (484.6,Wine)
    (121.6,Air)
};\addlegendentry{SHAP}
\addplot coordinates {
    (3.74,Loan)
    (2.29,Lungs)
    (2.46,Wine)
    (2.76,Air)
};
\addlegendentry{LIME}
\end{axis}
\end{tikzpicture}
    \caption{Explaining single instance in Publically Available Datasets: SHAP vs ProToMEx.}
    \label{fig:public:lime_vs_shape_vs_protomex}
\end{figure}
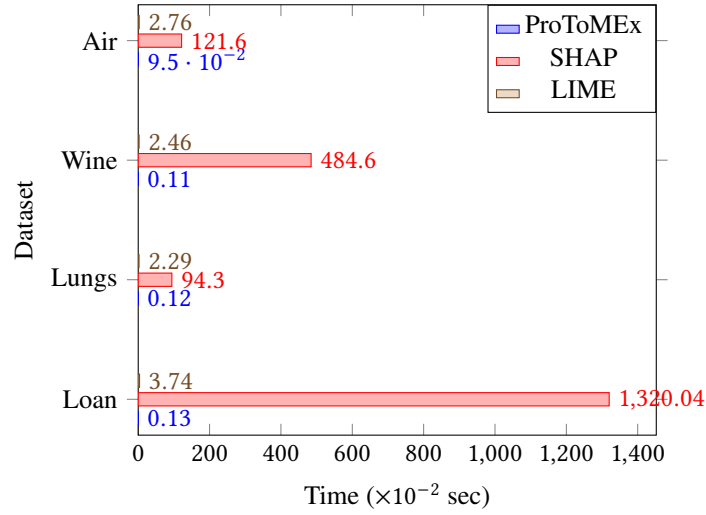
To evaluate the quality of our explainer, we consider the features appearing in the topics relevant to the instance at hand, and we compare them with the most influential ones according to SHAP and LIME.
Specifically, since both SHAP and LIME compute an importance score for each feature, we consider as most influential all the features that exceed $1/3$ of the average Shapley value and the average feature weight, respectively.
Then, we compute the features detected by ProToMEx, and consider the number of influential features that ProToMEx missed.
Tables~\ref{tab:quality:SHAP} and~\ref{tab:quality:LIME} hold the relevant information.
Specifically, $2^{nd}$ and $3^{rd}$ columns show the average number of influential features according to SHAP/LIME, and the average number of features detected by ProToMEx, respectively.
The remaining 10 columns show the average (over $100$ iterations) number of instances that missed $x = 0, \cdots, 10$ influential features per dataset.
As we can see, ProToMEx misses fewer than $3$ influential features (as identified either by SHAP or LIME) in the majority of the instances across all different datasets. 

Finally, we conduct an analysis to show that ProToMEx can identify influential features when SHAP fails.
Specifically, SHAP computes the Shapley value of each feature, which requires the consideration of all possible combinations of features.
Such a task is computationally demanding, and when the number of features increases, it becomes prohibitive. To overcome the computationally expensive nature of Shapley values, SHAP performs sampling of feature combinations. As such, in large datasets involving hundreds of features, SHAP might fail to sample the influential features.
To emulate this extreme situation, we force SHAP to ``miss'' sampling the top 3 influential features.
Since the Python API of SHAP does not allow us to directly hide some feature from sampling, we obfuscate the influential features by considering the mean value of this feature across the dataset.
Moreover, we reduce the number of samples used to $n=10$, while the default value is $n_\text{default} = 2 \cdot |F| + 2048$ with $|F|$ being the number of features.
In Table~\ref{tab:poorSHAP} we illustrate the average number of  missing influential features by the SHAP explainer, when SHAP is forced to miss the top-3 influential features---we refer to this explainer as `Poor Sampling SHAP'.
As we can see,  the Poor Sampling SHAP misses more than $6$ influential features in more instances than ProToMEx. Notably, the Wine and Air datasets, while ProToMEx in no instance misses more than $6$ features,  the Poor Sampling SHAP does in 9 and 3.37 instances.

Note that these results are not conclusive, however, they indicate that ProToMEx can identify influential features when SHAP misses sampling them. We expect this observation to become stronger when the number of features increases.
\begin{table}[H]
    \centering 
    \begin{tabular}{c|c|c|c|c|c}
    \multirow{2}{*}{Dataset}&\multicolumn{4}{|c}{Missing Influential Features}\\\cline{2-6}
         &$\geq6$&$\geq7$&$\geq8$&$\geq9$&$\geq10$  \\\hline
         Loan&5.38&2.88&1&0.13&0\\
         Lungs& 9.2& 5.08&3.04&2.08&1.32\\
         Wine&9&3.4&0.5&0&0\\
         Air&3.73&1.08&0.19&0.04&0\\
    \end{tabular}
    \caption{Average missing influential features by SHAP, when SHAP is forced to miss to sample the top-3 most influential features.}
    \label{tab:poorSHAP}
\end{table}


\vspace{-2mm}
\section{Conclusions}\label{sec:conclusion}
In this paper, we introduced ProToMEx, a novel general classifier explainer.
Our proposed explainer uses probabilistic topic modelling (PTM), and specifically LDA, to extract latent relations among the input features and the classifier's output.
We showed how to translate input data into documents to be fed to the PTM component, and we discussed how the topics can serve as global and local explanations.
Then we conducted a preliminary empirical evaluation to test the effectiveness of our approach.
We employed ProToMEx to explain classifiers trained with four different datasets, and we pitched it against the well-celebrated explainers SHAP and LIME.
Our results showed that we can provide explanations of similar quality with SHAP (for 75\% of the instances in three datasets misses less than 2 influential features).
At the same time, ProToMEx significantly outperforms SHAP in explaining single instances---it can yield an explanation at least $40\times$ faster.
Finally, we showed, ProToMEx manages to identify more influential features than a Poor Sample SHAP explainer that misses influential features.
This work is a practical step towards a new class of explainers that offer richer, more structured, and computationally scalable insights into complex models. The immediate direction for future research is to conduct user studies to formally evaluate the human-interpretability and utility of these topic-based explanations.

\begin{acks}
This research was supported by the Engineering and Physical Sciences Research Council (grant number EP/Y009800/1), through funding from Responsible Ai UK.
\end{acks}

\bibliographystyle{ACM-Reference-Format}
\bibliography{sample-base}

\end{document}